%% file: ZSFD_MM2023.tex
\pdfoutput=1
\documentclass[sigconf]{acmart}
\usepackage{balance}
\usepackage{amsmath}

\usepackage{amssymb}
\usepackage{amsthm}
\usepackage{bm}

\usepackage{booktabs} 
\usepackage{color}
\usepackage{algorithm}
\usepackage{algorithmic}
\usepackage{subfigure}
\usepackage{multirow}

\copyrightyear{2023}
\acmYear{2023}
\setcopyright{rightsretained}
\acmConference[MM '23]{Proceedings of the 31st ACM International Conference on Multimedia}{October 29-November 3, 2023}{Ottawa, ON, Canada}
\acmBooktitle{Proceedings of the 31st ACM International Conference on Multimedia (MM '23), October 29-November 3, 2023, Ottawa, ON, Canada}
\acmDOI{10.1145/3581783.3612661}
\acmISBN{979-8-4007-0108-5/23/10}

\begin{document}

\title{SeeDS: Semantic Separable Diffusion Synthesizer for Zero-shot Food Detection}

\author{Pengfei Zhou}
\affiliation{%
  \institution{Key Laboratory of Intelligent Information Processing, Institute of Computing Technology, Chinese Academy of Sciences (CAS)}
  \institution{University of Chinese Academy of Sciences}
  \city{Beijing}
  \country{China}
}
\email{pengfei.zhou@vipl.ict.ac.cn}

\author{Weiqing Min}
\authornote{Corresponding author.}
\authornote{Weiqing Min and Shuqiang Jiang are also with the Institute of Intelligent Computing Technology, Chinese Academy of Sciences, Suzhou, China.}
\affiliation{%
  \institution{Key Laboratory of Intelligent Information Processing, Institute of Computing Technology, CAS}
  \institution{University of Chinese Academy of Sciences}
  \city{Beijing}
  \country{China}
}
\email{minweiqing@ict.ac.cn}

\author{Yang Zhang}
\affiliation{%
  \institution{Key Laboratory of Intelligent Information Processing, Institute of Computing Technology, CAS}
  \institution{University of Chinese Academy of Sciences}
  \city{Beijing}
  \country{China}
}
\email{yang.zhang@vipl.ict.ac.cn}

\author{Jiajun Song}
\affiliation{%
  \institution{Key Laboratory of Intelligent Information Processing, Institute of Computing Technology, CAS}
  \institution{University of Chinese Academy of Sciences}
  \city{Beijing}
  \country{China}
}
\email{jiajun.song@vipl.ict.ac.cn}

\author{Ying Jin}
\affiliation{%
  \institution{Key Laboratory of Intelligent Information Processing, Institute of Computing Technology, CAS}
  \institution{University of Chinese Academy of Sciences}
  \city{Beijing}
  \country{China}
}
\email{yingyuan0226@gmail.com}

\author{Shuqiang Jiang$^\dag$}
\affiliation{%
  \institution{Key Laboratory of Intelligent Information Processing, Institute of Computing Technology, CAS}
  \institution{University of Chinese Academy of Sciences}
  \city{Beijing}
  \country{China}
}
\email{sqjiang@ict.ac.cn}

\renewcommand{\shortauthors}{Pengfei Zhou et al.}


\begin{abstract}
Food detection is becoming a fundamental task in food computing that supports various multimedia applications, including food recommendation and dietary monitoring. To deal with real-world scenarios, food detection needs to localize and recognize novel food objects that are not seen during training, demanding Zero-Shot Detection (ZSD). However, the complexity of semantic attributes and intra-class feature diversity poses challenges for ZSD methods in distinguishing fine-grained food classes. To tackle this, we propose the \textbf{Se}mantic S\textbf{e}parable \textbf{D}iffusion \textbf{S}ynthesizer (SeeDS) framework for Zero-Shot Food Detection (ZSFD). SeeDS consists of two modules: a Semantic Separable Synthesizing Module (S$^3$M) and a Region Feature Denoising Diffusion Model (RFDDM). The S$^3$M learns the disentangled semantic representation for complex food attributes from ingredients and cuisines, and synthesizes discriminative food features via enhanced semantic information. The RFDDM utilizes a novel diffusion model to generate diversified region features and enhances ZSFD via fine-grained synthesized features. Extensive experiments show the state-of-the-art ZSFD performance of our proposed method on two food datasets, ZSFooD and UECFOOD-256. Moreover, SeeDS also maintains effectiveness on general ZSD datasets, PASCAL VOC and MS COCO. The code and dataset can be found at \href{https://github.com/LanceZPF/SeeDS}{https://github.com/LanceZPF/SeeDS}.
\end{abstract}

\ccsdesc[500]{Computing methodologies~Object detection}

\keywords{food detection; zero-shot detection; food computing; zero-shot learning; diffusion model}


\maketitle

\input{Section1_Introduction_MM2023}

\input{Section2_Related_Work_MM2023}
\input{Section3_Method_MM2023}
\input{Section4_Experiment_MM2023}
\input{Section5_Conclusion_MM2023}

\begin{acks}
This work was supported by the National Nature Science Foundation of China (61972378, U19B2040, 62125207, U1936203), and was also sponsored by CAAI-Huawei MindSpore Open Fund.
\end{acks}

\bibliographystyle{ACM-Reference-Format}
\balance
\bibliography{ZSFD_MM2023}

\end{document}

%% file: Section1_Introduction_MM2023.tex
\section{Introduction}

\label{introduction}

\begin{figure}[t]
	\centering
	\includegraphics[width=8.2cm]{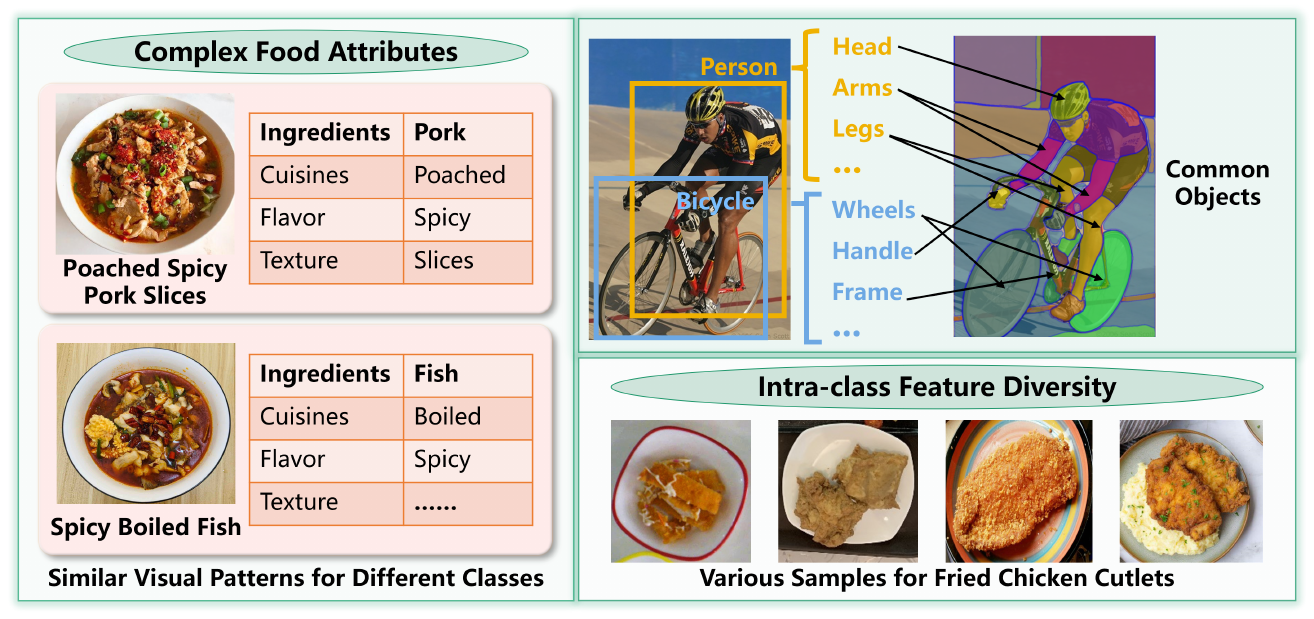}
	\caption{Our motivation: common objects possess certain semantic parts, and food objects lack such structural patterns to match with complex food attributes. Moreover, intra-class features diversify in the same food category.}
    \vspace{-0.6cm}
	\label{intro}
\end{figure}

Food computing~\cite{survey}, as an interdisciplinary field, utilizes computational methods to understand, model, and enhance human-food interactions, thereby offering a wide range of applications in health and nutrition areas ~\cite{Meyers-Im2Calories-ICCV2015, marin2019recipe1m, wang2022learning}. As one key task in food computing, food detection aims to locate and recognize food objects simultaneously \cite{aguilar2018grab, lu2020artificial, min2023large}. Food detection can enable various applications such as food recommendation, dietary assessment, and robotics control ~\cite{min2019food, wang2022review, ummadisingu2022cluttered}. However, it is challenging to detect food objects under real-world scenarios due to the constant emergence of novel food classes, such as the continued updates of food categories in restaurants \cite{shimoda2019webly}. In this case, continuously collecting and annotating new food objects is unrealistic. To address this, food detection needs the ability of Zero-Shot Detection (ZSD) to detect novel food classes that have no samples during training. 

ZSD emerges with the ability to detect unseen objects belonging to novel categories in real-world scenarios~\cite{bansal2018zero, zhu2020don}. It enables the transfer of knowledge from seen to unseen classes by incorporating semantic information from external sources like word embeddings. Based on this, ZSD researchers develop mapping-based methods \cite{berkan2018zero, yan2022semantics} and generation-based methods \cite{hayat2020synthesizing, huang2022robust}. The former learns a mapping function to align visual features and semantic features in a common space, tackling unknown objects by neighbor searching in this space. However, in a more complex setting of Generalized Zero-Shot Detection (GZSD)~\cite{bansal2018zero}, where both seen and unseen objects appear during inference, mapping-based methods would suffer from bias toward seen objects. To address this, generation-based methods are introduced with more robust GZSD performance. These methods use generative models (e.g., VAEs~\cite{kingma2013auto} and GANs~\cite{goodfellow2020generative}) to synthesize unseen features and train the zero-shot detector on these synthesized features. This paper reimplements general ZSD methods to food scenarios as an initial attempt at Zero-Shot Food Detection (ZSFD). However, current ZSD methods still meet difficulties when detecting unseen food categories since they are designed without food domain knowledge.

Compared with general ZSD, ZSFD is more challenging due to fine-grained issues of food categories, resulting in the complexity of food attributes and diversity of intra-class features. One of the major problems introduced by complex food attributes for ZSFD is semantic confusion. Similar to the concept of ZSD, ZSFD uses semantic information to bridge the gap between seen and unseen classes due to the lack of visual data for unseen objects. However, unlike common objects that often have distinct semantic parts for each class (e.g., head and limbs for \emph{Person}), food objects have no such structural visual patterns, making it harder to distinguish between different food objects with similar semantic attributes. For instance, \emph{Poached Spicy Pork Slices} and \emph{Spicy Boiled Fish} in Fig.~\ref{intro} share the same visual pattern, leading to difficulties distinguishing them using mere word embeddings. To address this semantic confusion, we need to extract multi-source semantic information, including ingredients and cuisines, to distinguish between similar food categories effectively.

Another challenge in ZSFD is the intra-class feature diversity, complicating fine-grained food detection. Specifically, dishes within the same food category can have completely different visual patterns. For example, instances of \emph{Fried Chicken Cutlets} have different appearances in Fig.~\ref{intro}, resulting in various visual features. Consequently, we need to synthesize diversified features for food classes to ensure the accuracy of zero-shot food detectors. However, existing generative models used in ZSD like GANs, have limitations in generating realistic and diversified food features for unseen food classes \cite{xiaotackling}. For example, the training of GANs is unstable and easily suffers from the model collapse problem since it is difficult to converge, which results in similar generated samples for each food class. Therefore, a stable generation model that can generate more diversified and realistic features for ZSFD tasks is to be explored.

To address these challenges, we propose Semantic Separable Diffusion Synthesizer (SeeDS), a novel ZSFD approach that overcomes these limitations by generating high-quality fine-grained features based on the advanced generative framework. SeeDS consists of two main modules: a Semantic Separable Synthesizing Module (S$^3$M) and a Region Feature Denoising Diffusion Model (RFDDM). The S$^3$M aims to enhance the semantic information by separating the food attributes according to two different domains: ingredient attributes and cuisine attributes. It can further learn disentangled semantic representation separately and synthesize discriminative food features for unseen classes via aggregating more abundant semantic information. The RFDDM leverages the latest diffusion model to generate food region features by reversing a Markov chain from noise to data. It takes the synthesized visual contents from the S$^3$M as the condition and applies a denoising process to generate more diverse and realistic food region features that can better capture the fine-grained characteristics of food items. Finally, a robust zero-shot food detector in SeeDS can be trained on the discriminative and diversified unseen food features.

Overall, our main contributions can be summarized as follows:
\begin{itemize}
\item We propose a novel Semantic Separable Diffusion Synthesizer (SeeDS), which overcomes the limitations of general ZSD frameworks by generating high-quality fine-grained features for detecting unseen food objects.

\item We introduce two modules in SeeDS. To address complex attribute issues, we present a Semantic Separable Synthesizing Module (S$^3$M) that enhances semantic information by learning disentangled ingredient and cuisine representation. To tackle intra-class feature diversity, we propose a Region Feature Denoising Diffusion Model (RFDDM) that leverages a novel diffusion model to generate more diverse and realistic food region features.

\item 
We evaluate our proposed framework on two food datasets ZSFooD and UECFOOD-256 \cite{kawano14c}, resulting in state-of-the-art ZSFD performance. Additionally, experiments on widely-used PASCAL VOC and MS COCO demonstrate the effectiveness of our approach for general ZSD. 

\end{itemize}

\begin{figure*}[t]
	\centering
    \includegraphics[width=16.8cm]{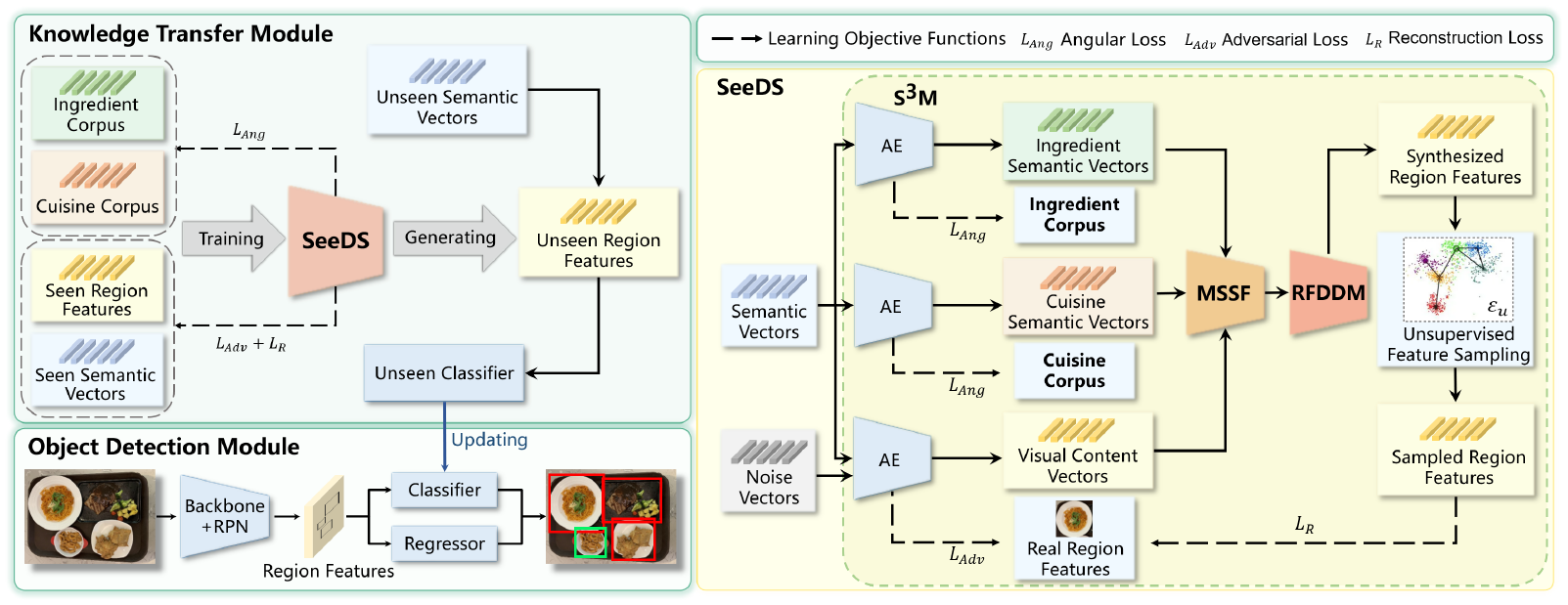}
	\caption{Framework of our approach. SeeDS consisting of the Semantic Separable Synthesizing Module (S$^3$M) and the Region Feature Denoising Diffusion Model (RFDDM) works as the knowledge transfer module. A zero-shot food detector can be obtained by combining an unseen classifier trained on the generated unseen features.} 
    \vspace{-0.4cm}
	\label{framework}
\end{figure*}

%% file: Section2_Related_Work_MM2023.tex
\section{Related Work}

\subsection{Zero-shot Learning}
Zero-Shot Learning (ZSL) is a machine learning branch enabling models to recognize unseen images~\cite{schonfeld2019generalized}. Existing ZSL methods utilize semantic information about novel images and primarily follow two zero-shot strategies: mapping-based approach \cite{socher2013zero} and generation-based approach \cite{schonfeld2019generalized}. Mapping-based approach projects extracted visual and semantic features into the same space and searches the nearest neighbor in the embedding space for input samples \cite{lei2015predicting}. Generation-based approach transforms zero-shot problems into supervised learning via synthesized features~\cite{gune2020generalized,han2021contrastive}. Specifically, generative models including VAEs~\cite{kingma2013auto} and GANs~\cite{goodfellow2020generative} are used to generate samples of novel classes based on their annotated attributes or semantic embeddings, which are obtained from large-scale pretrained language models like BERT~\cite{DevlinCLT19} and CLIP~\cite{radford2021learning}. Furthermore, the generated samples can then be used to train classifiers for recognizing novel unseen classes. 

\subsection{Zero-shot Detection}
Compared to the ZSL task, Zero-shot Detection (ZSD) presents greater challenges~\cite{tan2021survey, rahman2019zero}. ZSD approaches, including mapping-based~\cite{li2019zero, zheng2020background,yan2022semantics} and generation-based~\cite{zhao2020gtnet,hayat2020synthesizing,huang2022robust}, have been proposed, grounded in ZSL theory. For example, Bensal \textit{et al.}~\cite{bansal2018zero} propose two mapping-based approaches that use background-aware representations to improve the ZSD performance. They also provide a new evaluation metric called Generalized Zero-Shot Detection (GZSD), which aims to detect both seen and unseen classes during evaluation. Generation-based ZSD methods are developed with better GZSD performance~\cite{hayat2020synthesizing}. For example, Zhu \textit{et al.}~\cite{zhu2020don} propose an unseen feature generation framework based on VAE. Huang \textit{et al.}~\cite{huang2022robust} synthesize unseen features by a structure-aware GAN. However, due to the limitations of GANs and VAEs in generating diverse and realistic features, their real-world application proves difficult. Recently, diffusion models emerge as powerful generative models~\cite{ho2020denoising, songdenoising, rombach2022high}. Unlike GANs, they avoid training instability and model collapse and allow diversity control during generation. Therefore, we are the first to apply diffusion models to ZSD tasks.

\subsection{Food Detection}
Food detection \cite{aguilar2018grab} is an essential task in the field of food computing~\cite{survey}, attracting significant interest for its potential applications in the computer vision and multimedia community~\cite{aslan2020benchmarking, Ege2018Multi, rachakonda2020ilog}. However, food detection is a challenging task due to the complex characteristics of their ingredients, cuisines, flavors, etc. Additionally, it is difficult to distinguish between different food objects due to the intra-class variability and inter-class similarity of fine-grained food features~\cite{Ramdani2020Food}. One common approach for food detection is to implement general object detection frameworks. For example, Sun \textit{et al.}~\cite{sun2019foodtracker} propose a mobile application to detect food items based on YOLOv2~\cite{redmon2017yolo9000}. Shimoda \textit{et al.}~\cite{shimoda2019webly} propose a weakly-supervised food region proposal method using fully convolutional networks trained on web images. However, these food detection models meet difficulties when applied to real-world tasks. A major reason is that classes of meals are constantly updated in real-world scenarios (e.g., in restaurants), and food detectors trained with fixed classes can barely handle novel classes. Therefore, we introduce a novel zero-shot food detection framework SeeDS, which enables the effective zero-shot detection of novel food objects.

%% file: Section3_Method_MM2023.tex
\section{Method}

\noindent\textbf{Problem Formulation.}
The aim of the ZSFD task is to learn a detector on the training set $\mathcal{X}_s$ with semantic vectors and detect unseen objects in the test set. We define $\mathcal{X}_s$ that includes $A_s$ images, $B_s$ bounding box annotations, and $N_s$ seen food categories as the training set, and $\mathcal{O}_s$ as the available annotation set. Each food object $\bm{o}_m^{i} \in \mathcal{O}_s$ is annotated with a bounding box and a class label $y_m^{i} \in \mathcal{Y}_s$. $\mathcal{Y}_s = \{Y_1, ..., Y_{N_s}\}$ and $\mathcal{Y}_u = \{Y_{{N_s}+1}, ..., Y_N\}$ are class label sets of seen classes and unseen classes respectively, where $\mathcal{Y}_s \cap \mathcal{Y}_u = \emptyset$, $N = N_s + N_u$ is the number of all classes and $N_u$ is the number of unseen classes. Corresponding to class labels, semantic vector set $\mathcal{V} = \mathcal{V}_s \cup \mathcal{V}_u$ is given, where $\mathcal{V}_s$ and $\mathcal{V}_u$ are semantic vector sets of seen and unseen classes, respectively. The semantic vector $\bm{v} \in \mathcal{V}$ is word embeddings extracted from language models. During the inference, a test set $\mathcal{X}_t$ that contains both $N_s$ seen classes and $N_u$ unseen classes is given. ZSFD also evaluates methods on an unseen set $\mathcal{X}_u\subset \mathcal{X}_t$ that only contains unseen classes.

\noindent\textbf{Framework Overview.}
As shown on the left of Fig. \ref{framework}, the proposed framework consists of two parts. An object detection module $\phi_d$ based on the backbone detector is first trained with images in $\mathcal{X}_s$ containing food object annotations of seen classes, and then used to extract region features $\mathcal{E}_s$ of seen food objects. In the knowledge transfer module, we train a Semantic Separable Diffusion Synthesizer (SeeDS) $\bm{G}$ utilizing the extracted region features $\mathcal{E}_s$, the semantic vectors $\mathcal{V}_s$ according to their food classes, and corpora of ingredient and cuisine. Furthermore, we use $\bm{G}$ for generating robust and diverse unseen features $\mathcal{E}_u$ via unseen semantic vectors $\mathcal{V}_{u}$. An unseen classier $\phi_{uc}$ is further trained on the generated features $\mathcal{E}_u$ and combined into the original detector $\phi_d$. Updating the parameters in the detector with the parameters of the unseen classifier, an efficient zero-shot food detector that can locate and recognize unseen food objects is obtained. The framework of our approach is also summarized in Algorithm \ref{alg}.


\subsection{Semantic Separable Synthesizing Module}

\begin{algorithm}[!t]
\renewcommand{\arraystretch}{0.75}
\caption{The framework of our ZSFD approach}
\begin{algorithmic}[1] 
\REQUIRE Training set $\mathcal{X}_s$ with food images and annotations, seen semantic vector set $\mathcal{V}_s$ and unseen semantic vector set $\mathcal{V}_u$ 
\ENSURE Zero-shot food detector with parameters $\phi_d$ 
\STATE $\phi_d$ ← Train detector on $\mathcal{X}_s$ with annotations 
\STATE $\mathcal{E}_s$ ← Extract $\mathcal{E}_s$ region features from $\mathcal{X}_s$ via $\phi_d$
\STATE $\bm{G}$ ← Train Semantic Separable Diffusion Synthesizer $\bm{G}$ on $\mathcal{E}_s$ with corresponding semantic vectors from $\mathcal{V}_{s}$
\STATE $\mathcal{E}_u$ ← Synthesize unseen region features using $\bm{G}$ and $\mathcal{V}_{u}$ 
\STATE $\phi_{uc}$ ← Train unseen classifier $\phi_{uc}$ using $\mathcal{E}_u$ with class labels
\STATE $\phi_d$ ← Update parameters in $\phi_d$ with $\phi_{uc}$ 
\RETURN $\phi_d$ 
\end{algorithmic} 
\label{alg}
\end{algorithm}

\noindent\textbf{Disentangled Semantic Knowledge Learning.} 
In our proposed SeeDS, the Semantic Separable Synthesizing Module (S$^3$M) first learns the semantic representation of ingredients and cuisines based on Disentangled Semantic Knowledge Learning. To learn the semantic representation for fine-grained food classes with domain knowledge, we adopt a disentangled framework consisting of three branches. Two branches of these correspond to the semantic information of ingredients and cuisines. On each branch, an Auto-Encoder (AE) takes the word embeddings as the input semantic vectors, encodes them into latent semantic vectors, and decodes them into reconstructed semantic vectors.

We introduce two domain-knowledge corpora including an ingredient corpus and a cuisine corpus. Each corpus contains a bag of words that are relevant to the specific domain. For example, the ingredient corpus contains ingredient words like \emph{Tomato}, \emph{Eggs}, and \emph{Onion}, while the cuisine corpus contains cuisine words like \emph{Scrambled}, \emph{Stewed}, and \emph{Fried}. The objective of each branch in the disentangle framework is to minimize the angular loss between the input vector and the specific domain knowledge corpus. We also construct two learnable attention masks on both branches: an ingredient attention mask and a cuisine attention mask. Each attention mask is learned as a binary vector representing which words in the corpus are close to the class embedding vector. For example, if the class embedding vector is from \emph{Scrambled Eggs with Onion}, then the ingredient attention mask is learned to be [0, 1, 1] and the cuisine attention mask is learned to be [1, 0, 0]. We apply these learnable attention masks as the objective vectors $\bm{M}^p \in \mathbb{R}^{n \times a^p}$ in the calculation of the training objective on the $p$-th branch:

\begin{equation} 
\begin{aligned} 
\mathcal{L}_{Ang}^p =& -\frac{1}{n \cdot a_p} \sum_{i=1}^{n} \sum_{j=1}^{a^p} M^p_{i,j} \cdot  \log(\text{Sig}(\text{Ang}(\bm{\tilde{V}}_i, \bm{K}^p_{j}))) 
\\ &+ (1 - M^p_{i,j}) \cdot \log(1-\text{Sig}(\text{Ang}(\bm{\tilde{V}}_i, \bm{K}^p_{j}))) ,
\end{aligned}
\end{equation}
where $n$ is the batch size, $a^p$ is the size of a domain-knowledge corpus, $\bm{\tilde{V}}_i\in \mathbb{R}^{s}$ is the reconstructed semantic vector,  $\bm{K}^p_{j}\in \mathbb{R}^{s}$ is the $j$-th word vector in the $p$-th domain-knowledge corpus, $s$ is the dimension of word embeddings, $\text{Sig}(\cdot)$ is the sigmoid function, and $\text{Ang}(\bm{\tilde{V}}_i, \bm{K}^p_{j})$ is the cosine similarity between the decoded semantic vector and the corresponding corpus:

\begin{equation} 
\text{Ang}(\bm{\tilde{V}}_i, \bm{K}^p_{j}) = \dfrac {\bm{\tilde{V}}_i \cdot \bm{K}^p_{j}} {\max (\Vert \bm{\tilde{V}}_i \Vert _2 \cdot \Vert \bm{K}^p_{j} \Vert _2, \epsilon)},
\end{equation}
where $\epsilon$ is a small value to avoid division by zero. By optimizing the objective for decoded semantic embeddings and domain-knowledge corpora with learnable attention masks, two sets of semantic vectors are obtained: ingredient semantic vectors $\mathbf{V}_I \in\mathbb{R}^{n \times s}$ and cuisine semantic vectors $\mathbf{V}_C \in\mathbb{R}^{n \times s}$.

\noindent\textbf{Multi-Semantic Synthesis Fusion.}
The main challenge in our proposed S$^3$M is how to combine the separately generated semantic vector into a unified synthesized feature with rich knowledge. To address this challenge, we introduce Multi-Semantic Synthesis Fusion (MSSF) with a Content Encoder and a Fusion Decoder. 

\begin{figure}[t]
	\centering
	\includegraphics[width=7.8cm]{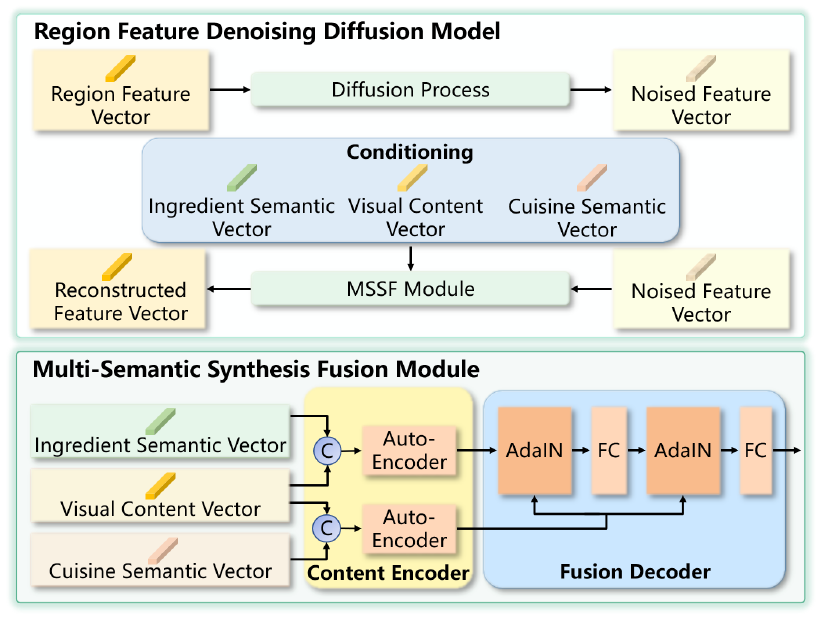}
	\caption{Detailed architecture of RFDDM and MSSF module. MSSF is used to predict the noise based on the conditions.}
	\label{fig:detail}
	\vspace{-0.5cm}
\end{figure}

As shown in Fig.~\ref{fig:detail}, an AE is applied as the Content Encoder for each branch. Each AE adopts two linear layers as the encoder $\text{ENC}(\cdot)$ and two linear layers as the decoder $\text{DEC}(\cdot)$, activated by LeakyReLU. For example, the Content Encoder takes $\mathbf{V}_I$ as input on the ingredient branch, encodes it with the visual content vectors $\bm{X}$ generated from the AE in the adversarial branch, and maps the latent representation into the ingredient content vectors:

\begin{align}
\bm{N}_I = \text{DEC}(\text{ENC}(\bm{Z} \otimes \bm{X} \otimes \bm{V}_I) \otimes \bm{X} \otimes \bm{V}_I),
\end{align}
where $\bm{X} \in \mathbb{R}^{n \times d}$, $\bm{Z} \in \mathbb{R}^{n \times d}$ are sampled noise vectors used for expanding the spanning space of synthesizing, $d$ is the dimension of the region feature, $\bm{V} \in \mathbb{R}^{n \times s}$ denotes input word embeddings, $\otimes$ denotes the concatenate operation, $\bm{N}_I \in \mathbb{R}^{n \times e}$ and $e$ is the new embedding dimension. The cuisine content vectors $\bm{N}_C \in \mathbb{R}^{n \times e}$ on the cuisine branch are obtained following the same pipeline.

The Fusion Decoder further takes separate content vectors as inputs and decodes them into synthesized features that combine content information from ingredients and cuisines. Two Adaptive Instance Normalization (AdaIN) \cite{huang2017arbitrary} blocks with two linear transformations are adopted to normalize the content with the semantic representation from the different branches:

\begin{align}
\text{AdaIN}(\bm{N}_I, \bm{N}_C) =  \sigma(\bm{N}_C) \left( \frac{\bm{\bm{N}_I}-\mu(\bm{\bm{N}_I})}{\sigma(\bm{N}_I)} \right) + \mu(\bm{N}_C),
\end{align}
where $\mu(\cdot)$ and $\sigma(\cdot)$ are the mean and standard deviation of vectors, respectively. Finally, we obtain synthesized features $\bm{E} \in \mathbb{R}^{b \times d}$, where $b$ is the synthesis number controlled by sample times.

\noindent\textbf{Unsupervised Feature Sampling.}
ZSFD needs high-quality features to learn a detector that can distinguish various unseen food objects. We should maximize the inter-class differentiation of synthesized features to deal with the fine-grained food problem. We use K-Means clustering to sample representative features $\bm{E}' \in \mathbb{R}^{b'\times d}$ from all generated data to balance the trade-off between the diversity and quality of generated samples and reduce the computational cost for further training. Specifically, we cluster synthesized features into $S$ clusters and select $b' = S \cdot P$ features from all clusters. We rank features in each cluster by their distance to the center. Assuming that the feature closer to the cluster center has higher quality and better generalization ability, we choose the top $P$ features according to their distance score in each cluster. We ensure that selected samples are evenly chosen from each cluster.

\subsection{Region Feature Denoising Diffusion Model}
The core generator in our SeeDS is a newly proposed Region Feature Denoising Diffusion Model (RFDDM). RFDDM can be used to generate 1D feature vectors, which can improve the diversity of synthesized region features in SeeDS. As shown in Fig.~\ref{fig:detail}, the RFDDM is based on the idea of modeling the data distribution learns to generate samples by applying a series of denoising steps to reverse the diffusion process that removes the sampled noise and recovers the region feature.

\begin{figure}[t]
	\centering
	\includegraphics[width=8cm]{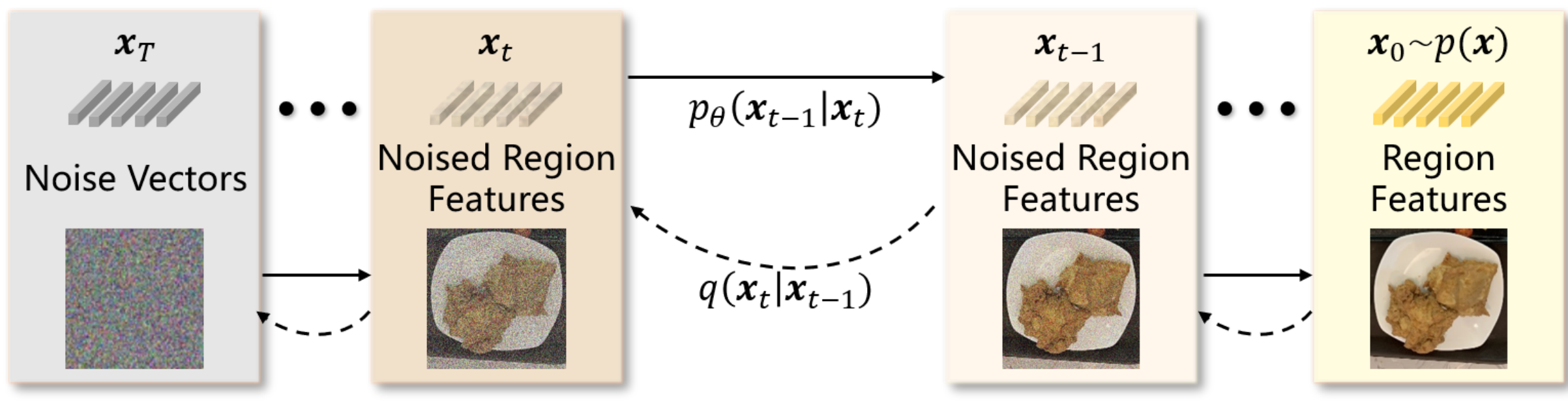}
	\caption{The visual illustration of the diffusion process in RFDDM. The forward process $q(\bm{x}_t|\bm{x}_{t-1})$ continually add Gaussian noise to $\bm{x}_{t-1}$ (from right to left), the reverse process $p_\theta(\bm{x}_{t-1}|\bm{x}_t)$ aims to denoise the noised feature vector $\bm{x}_{t}$.}
	\label{fig:dif}
	\vspace{-0.5cm}
\end{figure}

Let $\bm{x} \in \mathbb{R}^{d}$ be a 1D region feature vector. As illustrated in ~\ref{fig:dif}, we assume that $\bm{x}_T$ is generated by a diffusion process that starts from the sample $\bm{x}_0 \sim p_0(\bm{x})$, where $p_0$ is the data distribution, and Gaussian noise is added at each timestep $t = 1, …, T$ according to the Markovian process. The noise level at each timestep is controlled by a scalar $\beta_t \in (0, 1)$. The forward diffusion process $q(\bm{x}_t|\bm{x}_{t-1})$ for each timestep can be described as:

\begin{equation}
\bm{x}_t = \sqrt{1 - \beta_t} \bm{x}_{t-1} + \sqrt{\beta_t} \bm{z}_t ,
\end{equation}
where $\bm{z}_t  \in \mathbb{R}^{d}$ is sampled from $\mathcal{N}(\mathbf{0},\mathbf{I})$ and $\bm{x}_0 = \bm{x}$. The RFDDM aims to reverse this process, which is given by:

\begin{equation} 
p_{\theta}(\bm{x}_{t-1}|\bm{x}_t) = N (x_{t-1}; \mu_{\theta}(\bm{x}_t, t), \Sigma_{\theta}(\bm{x}_t, t)) .
\end{equation}

RFDDM uses the parameters of the MSSF module, which is shown in Fig.~\ref{fig:detail}, to predict the recover region feature utilizing the covariance $\Sigma_{\theta}(\bm{x}_t, t)$ and the mean $\mu_{\theta}(\bm{x}_t, t)$:

\begin{equation}
    \bm{\mu}_{\theta}(\bm{x}_t, t) = \frac {1} {\sqrt{\alpha_t}} ( \bm{x}_t - \frac{1-\alpha}{\sqrt{ 1 - \bar{\alpha}_{t} }} \bm{z}_\theta (\bm{x}_t, t) ),
\end{equation}
where $\alpha_t = 1 - \beta_t$, $\bar{\alpha}_t = \prod^t_{i=1} \alpha_i$ is the accumulated noise scalars, and $\bm{z}_\theta(\bm{x}_t, t)$ is the predicted noise parameterized by RFDDM. Thus we can map $\bm{x}_t$ to $\bm{x}_0$ by applying a series of denoising functions $\bm{F}_t$:
\begin{equation}
\bm{x}_{t-1} = \bm{F}_t(\bm{x}_t, t, \bm{z}_\theta(\bm{x}_t, t); \theta) ,
\end{equation}
where $\theta$ are the parameters of the MSSF module in RFDDM. The denoising functions $\bm{F}_t$ are implemented by MSSF modules that share the same architecture but have different parameters for each timestep. The RFDDM is trained by minimizing the mean squared error between the real noise $\bm{z}_t$ and $\bm{z}_\theta(\bm{x}_t, t)$ for all timesteps:

\begin{equation}
\begin{aligned}
\mathcal{L}_R &= \mathbb{E}_{\bm{x},\bm{z}_t}[\sum_{t=1}^T ||\bm{z}_{t} - \bm{z}_\theta(\bm{x}_t, t)||^2] \\
&=  \mathbb{E}_{\bm{x},\bm{z}}[\sum_{t=1}^T ||\bm{z}_{t} -  \bm{F}_t(\bm{x}_t, t, \bm{z}_\theta(\bm{x}_t, t); \theta)||^2] .
\end{aligned}
\end{equation}

\subsection{Loss Functions}
Given the seen feature collection $\mathcal{E}_s$ with semantic vector set $\mathcal{V}$ from $\mathcal{X}_s$ and Gaussian noise set $\mathcal{Z}$, our goal is to learn a synthesizer $\bm{G}$: $(\mathcal{V} \times \mathcal{Z}) \mapsto \mathcal{E}$, which takes a semantic vector $\bm{v} \in \mathcal{V}_u$ and a sampled random noise $\bm{z} \sim \mathcal{N}(0,\bm{I})$ as inputs and outputs the synthesized region feature $\tilde{\bm{e}} \in \mathcal{E}_u$. Specifically, the total training objective in SeeDS comprises three parts: the angular loss $\mathcal{L}_{Ang}^p$ that is used to learn domain-specific semantic representations, the adversarial loss $\mathcal{L}_{Adv}$ that is used to learn the visual content and the reconstruction loss $\mathcal{L}_R$ that is used to train the RFDDM.

\begin{align}
	\mathcal{L}_{Total} = \lambda_{1} (\mathcal{L}_{Ang}^1 + \mathcal{L}_{Ang}^2) + \lambda_{2} \mathcal{L}_{Adv} + \lambda_3 \mathcal{L}_{R}
 ,
\end{align}
where $\mathcal{L}_{Ang}^1$ and $\mathcal{L}_{Ang}^2$ are angular losses on ingredient and cuisine branches respectively. $\lambda_1$, $\lambda_{2}$ and $\lambda_3$ control the weights of the three losses. Specifically, the proposed $\mathcal{L}_{Adv}$ is used to adversarially train an AE that generates the visual content, which is one of the condition vectors for RFDDM to synthesize diverse samples:

\begin{align}
	\mathcal{L}_{Adv} = \mathop{\min}_{G} \mathop{\max}_{D} & \mathcal{L}_{W} + \mathcal{L}_{C} + \mathcal{L}_{S},
\end{align}
where $\mathcal{L}_{W}$ is the conditional Wasserstein GAN loss ~\cite{arjovsky2017wasserstein}, $\mathcal{L}_C$ is the classifier alignment loss~\cite{hayat2020synthesizing}, $\mathcal{L}_S$ is the semantic diverging loss~\cite{huang2022robust}.

%% file: Section4_Experiment_MM2023.tex
\section{Experiments}
\label{sec:experiments}


\begin{table}[!t]
  \centering
  \renewcommand{\arraystretch}{0.75}
  \caption{Statistics of the proposed ZSFooD and existing datasets that are widely used for ZSD.}
  	\setlength{\tabcolsep}{1.1mm}{
    \begin{tabular}{lccccccc}
    \toprule
    \multirow{2}[4]{*}{Dataset} & \multicolumn{3}{c}{Classes} & \multicolumn{3}{c}{Images} \\
\cmidrule{2-4}   \cmidrule{5-7}          & S     & U     & Total & Training & Test  & Total \\
    \midrule
    PASCAL VOC \cite{everingham2010pascal}   & 16    & 4     & 20    & 10,728 & 10,834 & 21,562 \\
    MS COCO \cite{coco} & 65    & 15    & 80    & 82,783 & 40,504 & 123,287 \\
    UECFOOD-256 \cite{kawano14c}  & 205    & 51     & 256    & 20,452 & 5,732 & 26,184 \\
    ZSFooD  & 184 & 44 & 228 & 10,463 & 10,140 & 20,603 \\
    \bottomrule
    \end{tabular}}
  \label{tab:dataset_comparison}%
\end{table}%

\begin{table}[!t]
  \centering
  \renewcommand{\arraystretch}{0.75}
  \caption{Comparison with baseline methods on ZSFooD (\%).}
    \setlength{\tabcolsep}{1.5mm}{
    \begin{tabular}{clcccc}
    \toprule
    \multirow{2}[4]{*}{Metric} & \multirow{2}[4]{*}{Method} & \multirow{2}[4]{*}{ZSD} & \multicolumn{3}{c}{GZSD} \\
\cmidrule{4-6}          &       &       & Seen  & Unseen & HM \\
    \midrule
   \multirow{6}[2]{*}{Recall@100} & ConSE~\cite{norouzi2014zero} & 39.7 & 58.0 & 38.1 & 46.4 \\
          & BLC~\cite{zheng2020background}  & 41.2  & 55.3   & 40.5  & 46.8 \\
          & CZSD~\cite{yan2022semantics}  &   48.0    & 86.1  &  44.8 &  58.9 \\
   & SU~\cite{hayat2020synthesizing}    & 45.3  & 82.3  & 44.1  & 57.4 \\
          & RRFS~\cite{huang2022robust}  & 48.8  & 86.6  & 47.6  & 61.4 \\
          & \textbf{SeeDS}  & \textbf{52.9}  & \textbf{87.0}  & \textbf{49.8}  & \textbf{63.3} \\
    \midrule
   \multirow{6}[2]{*}{mAP} & ConSE~\cite{norouzi2014zero} & 0.8 & 54.3 & 0.7 & 1.4 \\
          & BLC~\cite{zheng2020background}  & 1.1   & 51.1   & 0.9   & 1.8  \\
      & CZSD~\cite{yan2022semantics}  &  4.0   &    81.2   &   2.1  & 4.1 \\
       & SU~\cite{hayat2020synthesizing}    & 3.9   & 79.1  & 2.3   & 4.5  \\
          & RRFS~\cite{huang2022robust}  & 4.3   & 82.7  & 2.7   & 5.2  \\
          & \textbf{SeeDS}  & \textbf{5.9}   & \textbf{82.8}  & \textbf{3.5} & \textbf{6.7}  \\
    \bottomrule
    \end{tabular}}
  \label{tab:ZSFooD}%
\end{table}%

\subsection{Experimental Settings}

\noindent\textbf{Dataset Splittings.}
We adopt two datasets to evaluate the ZSFD performance: our constructed ZSFooD and widely-used UECFOOD-256 \cite{kawano14c}. UECFOOD-256 is a food detection dataset reformable into a ZSFD benchmark but mainly contains images with a single food object, and thus only provides 28,429 bounding box annotations. Compared with it, ZSFooD has 20,603 food images collected in 10 restaurant scenarios, each with multiple food objects annotated with bounding boxes. ZSFooD is more challenging with 95,322 bounding boxes and 291 classes. 
Following the setting in \cite{bansal2018zero, rahman2018polarity}, categories in ZSFooD and UECFOOD-256 are split into 184 seen classes and 44 unseen classes, and 205 seen classes and 51 unseen classes, respectively. We also compare our method with ZSD baselines on PASCAL VOC 2007+2012~\cite{everingham2010pascal} and MS COCO 2014 \cite{coco} using the given splitting \cite{hayat2020synthesizing, huang2022robust}. Note that two different splits are adopted for MS COCO: 48/17 seen/unseen split and 65/15 seen/unseen split. We replace the ingredient corpus with the corpus of texture and color words and replace the cuisine corpus with the corpus of shape and edge words when implementing SeeDS for general ZSD.

\begin{table}[!t]
  \centering
  \renewcommand{\arraystretch}{0.75}
  \caption{Comparison with baselines on UECFOOD-256 (\%).}
    \setlength{\tabcolsep}{1.5mm}{
    \begin{tabular}{clcccc}
    \toprule
    \multirow{2}[4]{*}{Metric} & \multirow{2}[4]{*}{Method} & \multirow{2}[4]{*}{ZSD} & \multicolumn{3}{c}{GZSD} \\
\cmidrule{4-6}          &       &       & Seen  & Unseen & HM \\
    \midrule
   \multirow{4}[2]{*}{Recall@100} 
          & CZSD~\cite{yan2022semantics}  &  60.7   & \textbf{57.6}  &  45.5 & 50.8 \\
   & SU~\cite{hayat2020synthesizing}    & 61.9 & 52.5  & 52.8  & 52.6 \\
          & RRFS~\cite{huang2022robust}  &  64.8 & 54.9  & 55.1  &  55.0 \\
          & \textbf{SeeDS}  & \textbf{74.0}  & 55.2  & \textbf{61.4}  & \textbf{58.1} \\
    \midrule
   \multirow{4}[2]{*}{mAP} 
      & CZSD~\cite{yan2022semantics}  &  22.0  &   \textbf{20.8}   &   16.2  & 18.2 \\
       & SU~\cite{hayat2020synthesizing}    &  22.4   &  19.3 & 20.1  & 19.7 \\
          & RRFS~\cite{huang2022robust}  &  23.6  & 20.1  &  22.9  &  21.4 \\
          & \textbf{SeeDS}  & \textbf{27.1}   & 20.2 & \textbf{26.0} & \textbf{22.7}  \\
    \bottomrule
    \end{tabular}}
  \label{tab:UEC256}%
\end{table}%

\noindent\textbf{Evaluation Metrics.}
Similar to previous works \cite{bansal2018zero, huang2022robust}, we use mean Average Precision (mAP) and Recall@100 with IoU threshold 0.5 for the evaluation on ZSFooD, UECFOOD-256 and PASCAL VOC. For MS COCO, we report mAP and Recall@100 with IoU thresholds of 0.4, 0.5, and 0.6. We also report the performance of methods under the setting of GZSD. The Harmonic Mean (HM) of seen and unseen is the key metric used for GZSD performance. 

\noindent\textbf{Implementation Details.}
We adopt the Faster-RCNN \cite{fasterrcnn} with the ResNet-101 \cite{resnet} as the backbone for fair comparisons. For training the synthesizer, Adam \cite{adam2014} is used with a learning rate of 1e-4 with a weight decay of 1e-5 for all experiments. To align the experimental settings with baselines \cite{hayat2020synthesizing,huang2022robust}, we synthesize 500/500/500/250 features for each unseen class of ZSFooD/UECFOOD-256/PASCAL VOC/MS COCO to train the classifier. We set $T = 100$ for the noise sampling process of RFDDM. The linear start and the linear end for noise scalars are set to $\beta_1$ = 8.5e-4 and $\beta_T$ = 1.2e-2, respectively. We empirically set $\lambda_1 = 1$, $\lambda_2 = 1$ and $\lambda_3 = 0.1$ without meticulous tuning for ensuring stable training. Word embedding vectors of class names and corpora are extracted by CLIP \cite{radford2021learning} for ZSFooD and UECFOOD-256, and extracted by FastText \cite{mikolov2018advances} for PASCAL VOC and MS COCO.

\begin{table}[!t]
  \centering
  \renewcommand{\arraystretch}{0.75}
  \caption{Comparison of mAP on PASCAL VOC (\%).}
    \setlength{\tabcolsep}{3.8mm}{
    \begin{tabular}{lcccc}
    \toprule
    \multirow{2}[4]{*}{Model} & \multirow{2}[4]{*}{ZSD} & \multicolumn{3}{c}{GZSD} \\
\cmidrule{3-5}          &       & Seen  & Unseen & HM \\
    \midrule
    SAN~\cite{rahman2019zero}   & 59.1  & 48.0    & 37.0    & 41.8 \\
    HRE~\cite{berkan2018zero}   & 54.2  & \textbf{62.4}  & 25.5  & 36.2 \\
    PL~\cite{rahman2018polarity}    & 62.1  & -     & -     & - \\
    BLC~\cite{zheng2020background}   & 55.2  & 58.2  & 22.9  & 32.9 \\
    SU~\cite{hayat2020synthesizing}    & 64.9  & -     & -     & - \\
    RRFS~\cite{huang2022robust}  & 65.5  & 47.1  & 49.1  & 48.1 \\
    \textbf{SeeDS}  & \textbf{68.9} & 48.5  & \textbf{50.6} & \textbf{49.5} \\
    \bottomrule
    \end{tabular}}
  \label{tab:voc}%
\end{table}%

\begin{table}[t]
  \centering
  \renewcommand{\arraystretch}{0.75}
  \caption{Comparison of Class-wise AP and mAP for different methods on unseen classes of PASCAL VOC (\%).}
    \setlength{\tabcolsep}{2.5mm}{
    \begin{tabular}{lccccc}
    \toprule
    Method & car   & dog   & sofa  & train & mAP \\
    \midrule
    SAN~\cite{rahman2019zero} & 56.2  & 85.3  & 62.6  & 26.4  & 57.6  \\
    HRE~\cite{berkan2018zero} & 55.0  & 82.0  & 55.0  & 26.0  & 54.5  \\
    PL~\cite{rahman2018polarity} & \textbf{63.7}  & 87.2  & 53.2  & 44.1  & 62.1  \\
    BLC~\cite{zheng2020background} & 43.7  & 86.0  & 60.8  & 30.1  & 55.2  \\
    SU~\cite{hayat2020synthesizing} & 59.6  & 92.7  & 62.3  & 45.2  & 64.9  \\
    RRFS~\cite{huang2022robust} & 60.1  & 93.0  & 59.7 & 49.1  & 65.5  \\
    \textbf{SeeDS}  & 60.4 & \textbf{95.3} & \textbf{65.9}  & \textbf{53.8} & \textbf{68.9} \\
    \bottomrule
    \end{tabular}}
  \label{tab:vocclass}%
\end{table}%

\begin{table}[t]
  \centering
  \renewcommand{\arraystretch}{0.75}
  \caption{ZSD performance comparison on MS COCO (\%).}
    \setlength{\tabcolsep}{0.9mm}{
    \begin{tabular}{lcrrrr}
    \toprule
    \multirow{2}[4]{*}{Model} & \multirow{2}[4]{*}{Split} & \multicolumn{3}{c}{Recall@100} & \multicolumn{1}{c}{mAP} \\
\cmidrule{3-6}          &       & \multicolumn{1}{c}{IoU=0.4} & \multicolumn{1}{c}{IoU=0.5} & \multicolumn{1}{c}{IoU=0.6} & \multicolumn{1}{c}{IoU=0.5} \\
    \midrule
    SB~\cite{bansal2018zero}    & 48/17 & \multicolumn{1}{c}{34.5} & \multicolumn{1}{c}{22.1} & \multicolumn{1}{c}{11.3} & \multicolumn{1}{c}{0.3} \\
    DSES~\cite{bansal2018zero}  & 48/17 & \multicolumn{1}{c}{40.2} & \multicolumn{1}{c}{27.2} & \multicolumn{1}{c}{13.6} & \multicolumn{1}{c}{0.5} \\
    PL~\cite{rahman2018polarity}    & 48/17 & \multicolumn{1}{c}{-} & \multicolumn{1}{c}{43.5} & \multicolumn{1}{c}{-} & \multicolumn{1}{c}{10.1} \\
    BLC~\cite{zheng2020background}   & 48/17 & \multicolumn{1}{c}{51.3} & \multicolumn{1}{c}{48.8} & \multicolumn{1}{c}{45.0} & \multicolumn{1}{c}{10.6} \\
    RRFS~\cite{huang2022robust}  & 48/17 & \multicolumn{1}{c}{58.1} & \multicolumn{1}{c}{53.5} & \multicolumn{1}{c}{47.9} & \multicolumn{1}{c}{13.4} \\
    \textbf{SeeDS}  & 48/17 &    \multicolumn{1}{c}{\textbf{59.2}}   & \multicolumn{1}{c}{\textbf{55.3}} &   \multicolumn{1}{c}{\textbf{48.5}}    & \multicolumn{1}{c}{\textbf{14.0}} \\
    \midrule
    PL~\cite{rahman2018polarity}    & 65/15 & \multicolumn{1}{c}{-} & \multicolumn{1}{c}{37.7} & \multicolumn{1}{c}{-} & \multicolumn{1}{c}{12.4} \\
    BLC~\cite{zheng2020background}   & 65/15 & \multicolumn{1}{c}{57.2} & \multicolumn{1}{c}{54.7} & \multicolumn{1}{c}{51.2} & \multicolumn{1}{c}{14.7} \\
    SU~\cite{hayat2020synthesizing}    & 65/15 & \multicolumn{1}{c}{54.4} & \multicolumn{1}{c}{54.0} & \multicolumn{1}{c}{47.0} & \multicolumn{1}{c}{19.0} \\
    RRFS~\cite{huang2022robust}  & 65/15 & \multicolumn{1}{c}{65.3} & \multicolumn{1}{c}{62.3} & \multicolumn{1}{c}{55.9} & \multicolumn{1}{c}{19.8} \\
    \textbf{SeeDS}  & 65/15 &    \multicolumn{1}{c}{\textbf{66.5}}   &   \multicolumn{1}{c}{\textbf{64.0}}    &    \multicolumn{1}{c}{\textbf{56.8}}   & \multicolumn{1}{c}{\textbf{20.6}} \\
    \bottomrule
    \end{tabular}}
  \label{tab:cocozsd}%
\end{table}%

\begin{table}[!t]
  \centering
  \renewcommand{\arraystretch}{0.75}
  \caption{GZSD performance comparison on MS COCO (\%).}
    \setlength{\tabcolsep}{1.2mm}{
    \begin{tabular}{lccccccc}
    \toprule
    \multirow{2}[4]{*}{Model} & \multirow{2}[4]{*}{Split} & \multicolumn{3}{c}{Recall@100} & \multicolumn{3}{c}{mAP} \\
\cmidrule{3-8}          &       & S     & U     & HM    & S     & U     & HM \\
    \midrule
    PL~\cite{rahman2018polarity}    & 48/17 & 38.2  & 26.3  & 31.2  & 35.9  & 4.1   & 7.4  \\
    BLC~\cite{zheng2020background}   & 48/17 & 57.6  & 46.4  & 51.4  & 42.1  & 4.5   & 8.1 \\
    RRFS~\cite{huang2022robust}  & 48/17 & 59.7  & 58.8  & 59.2  & 42.3  & 13.4  & 20.4  \\
    \textbf{SeeDS}  & 48/17 &  \textbf{60.1}     &   \textbf{60.8}   &    \textbf{60.5}   &    \textbf{42.5}  &     \textbf{14.5}  & \textbf{21.6} \\
    \midrule
    PL~\cite{rahman2018polarity}    & 65/15 & 36.4  & 37.2  & 36.8  & 34.1  & 12.4  & 18.2  \\
    BLC~\cite{zheng2020background}   & 65/15 & 56.4  & 51.7  & 53.9  & 36.0  & 13.1  & 19.2  \\
    SU~\cite{hayat2020synthesizing}    & 65/15 & 57.7  & 53.9  & 55.7  & 36.9  & 19.0  & 25.1  \\
    RRFS~\cite{huang2022robust}  & 65/15 & 58.6  & 61.8  & 60.2  & 37.4  & 19.8  & 26.0  \\
    \textbf{SeeDS}  & 65/15 &   \textbf{59.3}   &  \textbf{62.8}  &  \textbf{61.0}     &  \textbf{37.5}   &  \textbf{20.9}  &  \textbf{26.8} \\
    \bottomrule
    \end{tabular}}
  \label{tab:cocogzsd}%
\end{table}%

\subsection{Experiments on ZSFD datasets}
\label{sec:main_results}

\noindent\textbf{Evaluation on ZSFooD.}
We reimplement baseline methods and show ZSFD results on ZSFooD in Table~\ref{tab:ZSFooD}. Compared with the second-best method RRFS, ``ZSD'', ``Unseen'' and ``HM'' are improved by 1.6\%, 0.8\%, and 1.5\% mAP, respectively. For Recall@100, ``ZSD'', ``Unseen'' and ``HM'' are improved by 4.1\%, 2.2\% and 1.9\% mAP, respectively. The improvements demonstrate the effectiveness of the proposed SeeDS in detecting unseen fine-grained food objects. Our proposed SeeDS enhances feature synthesizer by utilizing the multi-source semantic knowledge from S$^3$M, and help detectors achieve better ZSFD performance when trained with diverse features generated by RFDDM. We also observe that the mAP performance of ``Unseen'' for all ZSD methods is much lower than mAP of ``Seen'', which denotes that the larger number of classes in ZSFooD makes ZSFD on unseen objects extremely challenging. Note that the ``Seen'' performance in the setting of GZSD has not been improved since classifier parameters for seen classes are mainly influenced by the backbone detector trained on seen objects, while ZSFD specifically focuses on improving detection on unseen classes.

\noindent\textbf{Evaluation on UECFOOD-256.} 
Experimental results on UECFOOD-256 are shown in Table~\ref{tab:UEC256}. Compared with RRFS, our SeeDS improves mAP by 3.5\%, 3.1\%, and 1.3\%, and Recall@100 by 9.2\%, 6.3\%, and 3.1\% for ``ZSD'', ``Unseen'' and ``HM'', respectively. For Recall@100, ``ZSD'', ``Unseen'' and ``HM'' are improved by 9.2\%, 6.3\%, and 3.1\%, respectively. These results underline the effectiveness of our SeeDS framework, especially in complex GZSD scenarios with simultaneous seen and unseen food objects. We also observe that the mAP for all baseline methods can not reach a similarly high number as in ZSD, indicating ZSFD as a challenging task with potential for further method development.

\subsection{Experiments on general ZSD datasets}

\begin{table}[t]
  \centering
  \renewcommand{\arraystretch}{0.75}
  \caption{Ablation studies measured by mAP (\%).}
    \setlength{\tabcolsep}{1.5mm}{
    \begin{tabular}{cccccccc}
    \toprule
    \multicolumn{1}{c}{\multirow{2}[4]{*}{Dataset}} & \multicolumn{2}{c}{Methods} & \multirow{2}[4]{*}{ZSD} & \multicolumn{3}{c}{GZSD} \\
\cmidrule{2-3}\cmidrule{5-7}          &  S$^3$M & RFDDM   &       & S     & U     & HM \\
    \midrule
    \multirow{3.5}[4]{*}{ZSFooD}  &   &    &  4.3   & 82.7  & 2.7   & 5.2   \\
      &    \checkmark  &  &   5.0   & 82.8   &  3.3 &  6.3 \\
       &   &  \checkmark  &  5.7  &  82.8 &  3.1 &  6.0 \\
       & \checkmark     & \checkmark     &  \textbf{5.9}   & \textbf{82.8}  & \textbf{3.5} & \textbf{6.7}  \\
    \midrule
    \multirow{3.5}[4]{*}{PASCAL VOC}      &       &        & 65.5  & 47.1  & 49.1  & 48.1 \\
       &   \checkmark     &  &    68.0   & 48.5 & 49.8 & 49.1 \\
        &     &    \checkmark    & 68.2 & 48.4 &  49.6 & 49.0 \\
      & \checkmark     & \checkmark     & \textbf{68.9} & \textbf{48.5}  & \textbf{50.6} & \textbf{49.5}  \\
    \bottomrule
    \end{tabular}}
  \label{tab:ablation}%
\end{table}%

\begin{figure}[t]
	\centering
	\includegraphics[width=7.8cm]{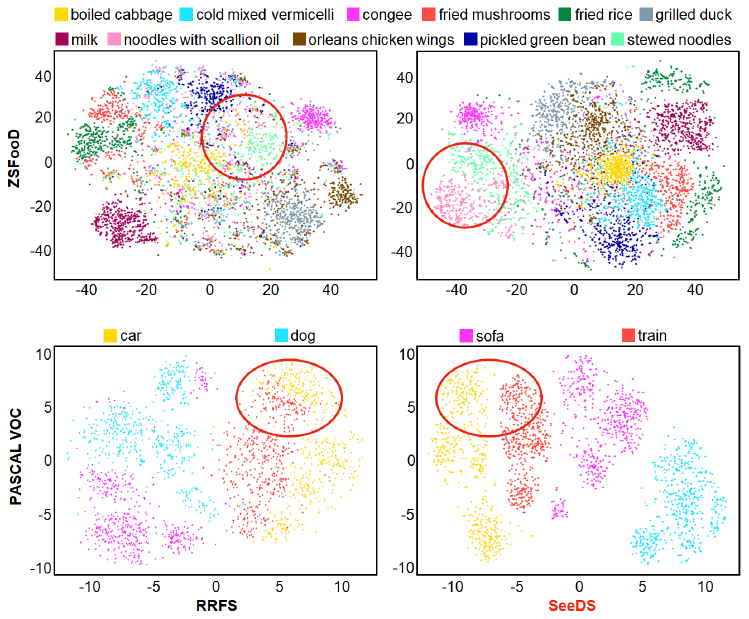}
	\caption{The t-SNE visualization of synthesized features.}
	\label{fig:label_tsne}
	\vspace{-0.35cm}
\end{figure}

\begin{figure*}[t]
	\centering
	\includegraphics[width=0.95\textwidth]{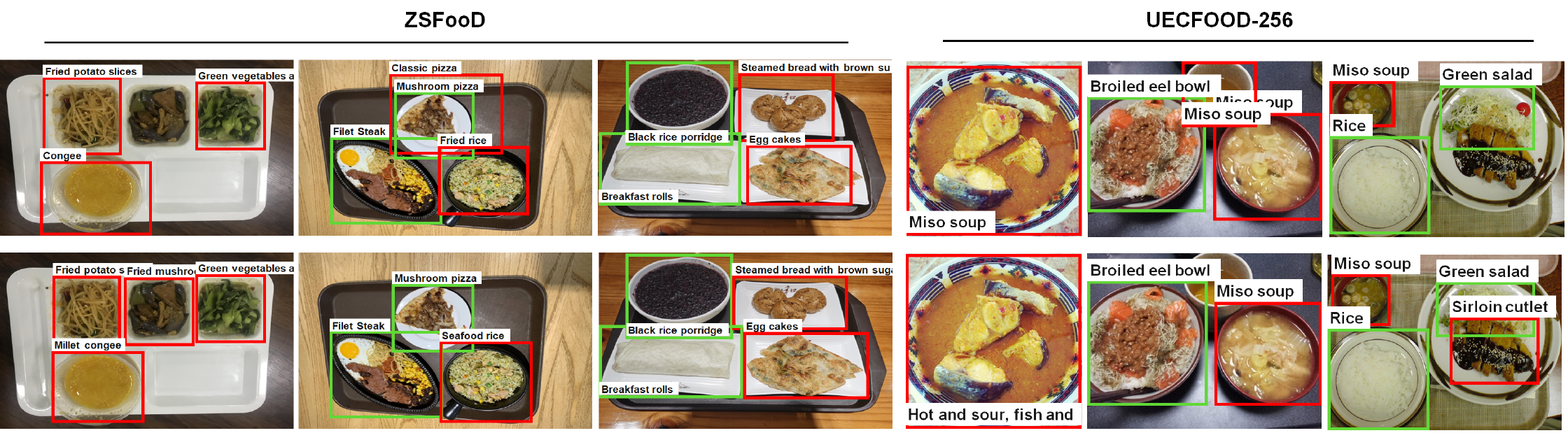}
	\caption{Detection results by baseline RRFS and our approach on ZSFooD, UECFOOD-256, PASCAL VOC and MS COCO. Seen classes are shown in green boxes and unseen in red boxes. Zoom in for a better experience.}
	\vspace{-0.3cm}
	\label{fig:qualitative}
\end{figure*}

\noindent\textbf{Evaluations on PASCAL VOC and MS COCO.}
To further evaluate the performance of SeeDS in general ZSD, we conduct extension experiments on PASCAL VOC and MS COCO. Our SeeDS outperforms all baselines under ``ZSD'' setting, increasing the mAP by 3.4\% compared with the latest ZSD baseline RRFS \cite{huang2022robust}. Furthermore, our method obtains better performance under a more challenging setting of GZSD. The ``Seen'', ``Unseen'' and ``HM'' are improved by 1.4\%, 1.5\% and 1.4\% compared with the RRFS. Results show that our method achieves a more balanced performance on the seen and unseen classes for GZSD. The class-wise AP performance on PASCAL VOC is reported in Table~\ref{tab:vocclass}. We can observe that our approach achieves the best performance in most classes.

We evaluate the ZSD performance on MS COCO with different IoU thresholds of 0.4, 0.5 and 0.6. As seen in Table \ref{tab:cocozsd}, our method outperforms all baseline methods, achieving significant gain on both mAP and Recall@100. For the 47/17 split, our method improves the mAP and Recall@100 by 0.6\% and by 1.8\% at IoU=0.5 compared with RRFS, respectively. For the 65/15 split, our SeeDS improves the mAP and Recall@100 by 0.8\% and by 1.7\% at IoU=0.5, respectively. As shown in Table \ref{tab:cocogzsd}, our SeeDS also outperforms the RRFS under the GZSD setting, where ``S'' denotes performance on seen classes and ``U'' denotes performance on unseen classes. The absolute ``HM'' performance gain of our method is 1.2\% mAP and 1.3\% Recall@100 for the 48/17 split, and 0.8\% mAP and 0.8\% Recall@100 for the 65/15 split. Results demonstrate that our model exceeds existing ZSD methods in terms of both mAP and Recall@100.

\subsection{Ablation Study}
\label{sec:ablation}

We conduct quantitative ablation analysis for two key modules, including the S$^3$M and RFDDM modules. Table \ref{tab:ablation} reports the ``ZSD'' and ``GZSD'' performance of mAP at IoU=0.5 for ZSFooD and PASCAL VOC. We replace the ingredient corpus with the texture corpus and replace the cuisine corpus with the shape corpus when implementing S$^3$M on PASCAL VOC. As shown in Table \ref{tab:ablation}, the SeeDS incorporating the strategy of S$^3$M improves the ``HM'' by 1.1\% on ZSFooD and 1.0\% on PASCAL VOC. This indicates that the synthesizer with richer semantic information can synthesize more robust visual features for ZSFD and ZSD. We can observe that the ``ZSD'' performance has been improved by 1.4\% on ZSFooD and 1.3\% on PASCAL VOC compared with the baseline when the core GAN-based synthesizer is replaced with the RFDDM. These performance gains demonstrate the effectiveness of implementing the diffusion model in the ZSD framework. With respect to the ``ZSD'', ``U'' and ``HM'', the 1.6\%, 0.8\% and 1.5\% mAP improvement on ZSFooD and the 3.5\%, 1.5\% and 1.4\% mAP improvement on PASCAL VOC are obtained by the SeeDS that adopts both S$^3$M and RFDDM modules compared with the baseline. It shows that all proposed modules are vital for providing more robust synthesized features for training an effective Zero-shot Detector, which is able to improve the ZSFD and ZSD performance by a large margin.

\subsection{Qualitative Results}
\label{sec:qualitative}

\noindent\textbf{Feature distribution visualization.}
To further demonstrate the effectiveness of our model in optimizing the distribution structure in generating, we utilize t-SNE \cite{van2008visualizing} to visualize generated unseen features on ZSFooD and PASCAL VOC. Region features generated by the baseline RRFS and our approach are illustrated in Figure \ref{fig:label_tsne}, where we select a quarter of the categories in ZSFooD to make visualization clear. Generated features for similar classes (e.g., \emph{Stewed Noodles} and \emph{Noodles with Scallion Oil} for ZSFooD, and \emph{Car} and \emph{Train} for PASCAL VOC) are confused with each other by the baseline method because of high similarity in their semantic representation. In this case, we observe that our synthesized features are more discriminative, which form well-separated clusters. Furthermore, discriminative synthesized features can help learn a more robust unseen classifier for ZSFD and ZSD.

\noindent\textbf{Detection Results.}
We visualize the results of ZSFD on ZSFooD and UECFOOD-256 in Figure~\ref{fig:qualitative}, where the first row is the output by RRFS and the second row is by our SeeDS. The baseline RRFS fails to predict true class labels for several unseen food objects, while our model provides more accurate ZSFD results. The proposed separable semantic learning in SeeDS effectively leverages domain knowledge of ingredients and cuisines to train an unseen classifier based on inter-class separable synthesized features. Furthermore, the incorporation of RFDDM further improves the performance of the synthesizer in generating fine-grained features that are robust for ZSFD. It is worth noting that ZSD baselines are often affected by similar visual features among fine-grained categories, even when their ingredients differ. For instance, the \emph{Hot and Sour, Fish and Vegetable Ragout} is mistakenly recognized as the \emph{Miso Soup} by RRFS. In contrast, SeeDS is able to discriminate between the \emph{Classic Pizza} and the \emph{Mushroom Pizza} via the difference in ingredients.

%% file: Section5_Conclusion_MM2023.tex
\section{Conclusion}

In this paper, we first define Zero-Shot Food Detection (ZSFD) task for tackling real-world problems. Furthermore, We propose a novel ZSFD framework SeeDS to address the challenges posed by complex attributes and diverse features of food. We evaluate our method and baselines on two food benchmark datasets ZSFooD and UECFOOD-256, which demonstrates the effectiveness and robustness of SeeDS on ZSFD. To further explore the performance of our method on general ZSD, we evaluate SeeDS on two widely-used datasets, PASCAL VOC and MS COCO. The results show that SeeDS can generalize well on ZSD. The ablation studies on ZSFooD and UECFOOD-256 demonstrate the effectiveness of the proposed modules, including S$^3$M and RFDDM. Therefore, our approach shows great potential for ZSFD and can be extended to various multimedia applications. In future research, we need to better understand the feature sampling mechanism and its potential for solving fine-grained problems. Also, with the development and new opportunities, we believe open-vocabulary food detection with a food-specialized visual-language pretrained model could better serve real-world needs and deserves further exploration.